\pdfoutput=1

\documentclass[11pt]{article}
\usepackage{graphicx}
\usepackage{acl}
\usepackage{times}
\usepackage{latexsym}
\usepackage{graphicx}
\usepackage{caption}
\usepackage{multicol}
\usepackage{multirow}
\usepackage{tabularx}
\usepackage{amsmath}
\usepackage{booktabs}
\usepackage{adjustbox}
\usepackage{footmisc}
\usepackage[misc]{ifsym}
\usepackage{subcaption}
\usepackage{extarrows}
\usepackage{amsmath}
\usepackage{mathrsfs}
\usepackage[utf8]{inputenc}
\usepackage{microtype}
\usepackage{color}
\usepackage{colortbl} 
\usepackage{mathrsfs}
\usepackage{float}
\usepackage[T1]{fontenc}

\usepackage[utf8]{inputenc}

\usepackage{microtype}

\title{RoMe: A Robust Metric for Evaluating Natural Language Generation}

 \author{Md Rashad Al Hasan Rony$^{1,3}$, Liubov Kovriguina$^{3}$, Debanjan Chaudhuri$^{1}$, \\\textbf{Ricardo Usbeck}$^{2}$\textbf{, Jens Lehmann}$^{1,3}$ \\ {$^{1}$University of Bonn, $^{2}$University of Hamburg, $^{3}$Fraunhofer IAIS Dresden} \\ \texttt{\{rashad.rony,liubov.kovriguina,jens.lehmann\}@iais.fraunhofer.de}\\\texttt{\{lehmann,d.chaudhuri\}@uni-bonn.de}\\\texttt{ricardo.usbeck@uni-hamburg.de}}

\begin{document}
\maketitle
\begin{abstract}
Evaluating Natural Language Generation (NLG) systems is a challenging task. Firstly, the metric should ensure that the generated hypothesis reflects the reference's semantics. Secondly, it should consider the grammatical quality of the generated sentence. Thirdly, it should be robust enough to handle various surface forms of the generated sentence. Thus, an effective evaluation metric has to be multifaceted. In this paper, we propose an automatic evaluation metric incorporating several core aspects of natural language understanding (language competence, syntactic and semantic variation). Our proposed metric, RoMe, is trained on language features such as semantic similarity combined with tree edit distance and grammatical acceptability, using a self-supervised neural network to assess the overall quality of the generated sentence. Moreover, we perform an extensive robustness analysis of the state-of-the-art methods and RoMe. Empirical results suggest that RoMe has a stronger correlation to human judgment over state-of-the-art metrics in evaluating system-generated sentences across several NLG tasks.
\end{abstract}

\section{Introduction}
Automatic generation of fluent and coherent natural language is a key step for human-computer interaction. 
Evaluating generative systems such as text summarization, dialogue systems, and machine translation is challenging since the assessment involves several criteria such as content determination, lexicalization, and surface realization~\cite{liu-etal-2016-evaluate,dale1998towards}. For assessing system-generated outputs, human judgment is considered to be the best approach. 
Obtaining human evaluation ratings, on the other hand, is both expensive and time-consuming. As a result, developing automated metrics for assessing the quality of machine-generated text has become an active area of research in NLP.

The quality estimation task primarily entails determining the similarity between the reference and hypothesis as well as assessing the hypothesis for grammatical correctness and naturalness.
%
Widely used evaluation metrics such as BLEU~\cite{papineni2002bleu}, METEOR~\cite{banerjee2005meteor}, and ROUGE~\cite{lin-2004-rouge} which compute the word-overlaps, were primarily designed for evaluating machine translation and text summarization systems. 
%
Word-overlap based metrics, on the other hand, are incapable of capturing the hypotheses' naturalness and fluency. Furthermore, they do not consider the syntactic difference between reference and hypothesis. In a different line of research, word mover distance (WMD)~\cite{kusner2015word}, BERTScore~\cite{zhang-2020BERTScore} and MoverScore~\cite{zhao-etal-2019-moverscore} compute word embedding based similarity for evaluating system-generated texts. Although these metrics employ the contextualized representation of words, they do not take the grammatical acceptability of the hypothesis and the syntactical similarity to the reference into account.

To address these shortcomings, we propose RoMe, an automatic and robust metric for evaluating NLG systems. 
RoMe employs a neural classifier that uses the generated sentence's grammatical, syntactic, and semantic qualities as features to estimate the quality of the sentence. \textbf{Firstly}, it calculates the earth mover's distance (EMD)~\cite{rubner1998metric} to determine how much the hypothesis differs from the reference. 
During the computation of EMD, we incorporate hard word alignment and soft-penalization constants to handle various surface forms of words in a sentence, such as repeated words and the passive form of a sentence. 
\textbf{Secondly}, 
using a semantically enhanced tree edit distance, the difference in syntactic structures between the reference and hypothesis sentences is quantified. \textbf{Thirdly}, the metric incorporates a binary classifier to evaluate the grammatical acceptability of the generated hypotheses. 
\textbf{Finally}, the scores obtained from the preceding steps are combined to form a representation vector, which is subsequently fed into a self-supervised network. 
The network produces a final score, referred to as RoMe's output which represents the overall quality of the hypothesis statement. 

We investigate the effectiveness of our proposed metric by conducting experiments on datasets from various domains of NLG such as knowledge graph based language generation dataset (KELM~\cite{agarwal-etal-2021-knowledge}), dialogue datasets~\cite{eric2017key,chaudhuri2021grounding}, the WebNLG 2017 challenge dataset~\cite{shimorina-2020-webnlg-human-eval}, structured data to language generation dataset (BAGEL~\cite{bagel} and SFHOTEL~\cite{wen-etal-2015-semantically}). The capability of existing metrics to handle various forms of text has lately become a matter of debate in the NLP community ~\cite{ribeiro-etal-2020-beyond,novikova-etal-2017-need,liu-etal-2016-evaluate}. Hence, we conduct an extensive robustness analysis to assess RoMe's performance in handling diverse forms of system-generated sentences. To verify our claim, we design the analysis based on the text perturbation methods used in CHECKLIST~\cite{ribeiro-etal-2020-beyond} and adversarial text transformation techniques from TextFooler~\cite{jin2020bert} and TextAttack~\cite{morris2020textattack}. Empirical assessment on benchmark datasets and the robustness analysis results exhibit that RoMe can handle various surface forms and generate an evaluation score, which highly correlates with human judgment. RoMe is designed to function at the sentence level and can be used to evaluate English sentences in the current version of the implementation. In the future versions, we plan to extend RoMe by including more languages. We released the code and annotation tool 
publicly~\footnote{\url{https://github.com/rashad101/RoMe}}.
 
    
    


\section{Preliminaries}
\subsection{Earth Mover's Distance}
The Earth Mover's Distance (EMD) estimates the amount of work required to transform a probability distribution into another~\cite{rubner1998metric}. Inspired by the EMD, in NLP the transportation problem is adopted to measure the amount of work required to match the system generated hypothesis sentence with the reference sentence~\cite{kusner2015word,zhao-etal-2019-moverscore}.
\begin{figure}[htb!]
\centering
\includegraphics[width=7.5cm]{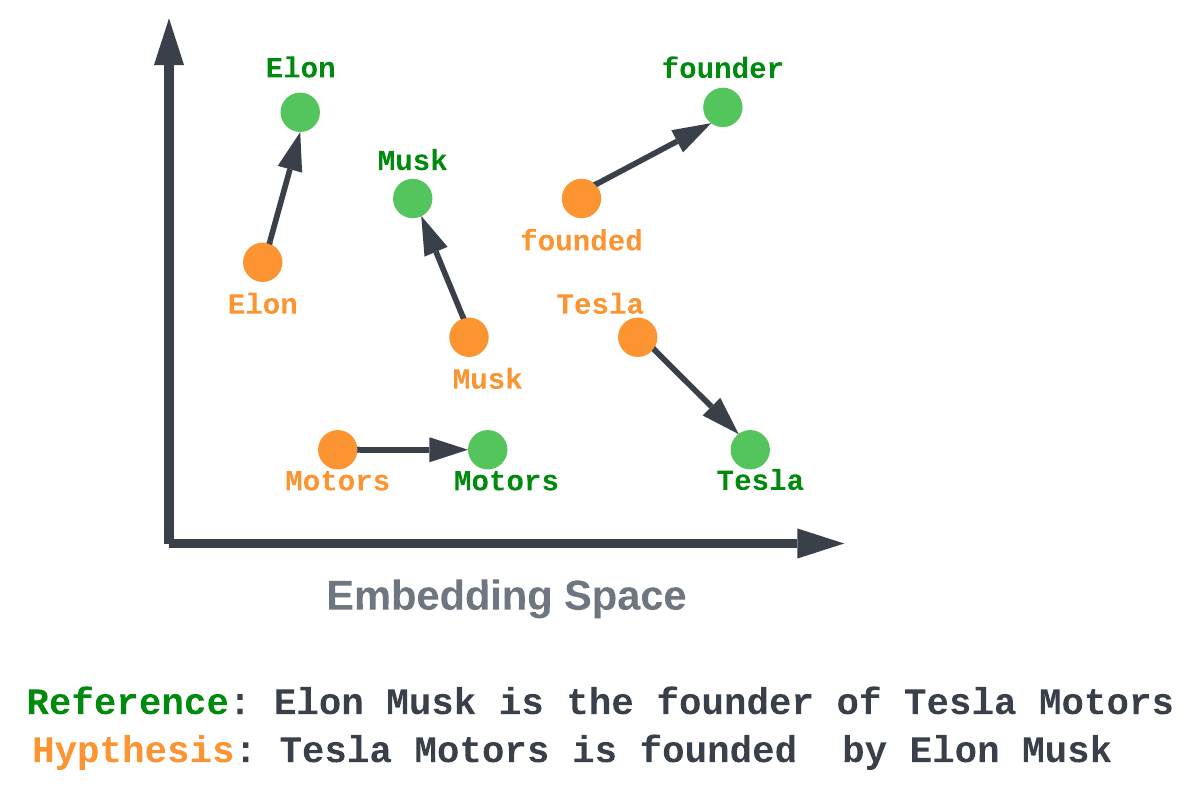}
    \caption{Illustrating an abstraction of the EMD.}
    \label{fig:emdintro}
\end{figure}
Let us define the reference as $\mathcal{R}$ = $\{r_{1},r_{2},...,r_{p}\}$ and the hypothesis as $\mathcal{H}$ = $\{h_{1},h_{2},...,h_{q}\}$, where $r_{i}$ and $h_{j}$ indicates the $i$-th and $j$-th word of the reference and hypothesis, respectively. The weight of the word $r_{i}$ and $h_{j}$ are denoted as $m_{i}$ and $n_{j}$ respectively. 
Then, the total weight distribution of $\mathcal{R}$ and $\mathcal{H}$ is $m_{\sum}$ = $\sum_{i=1}^{p} {m_{i}}$ and $n_{\sum}$ = $\sum_{j=1}^{q} n_{j}$, respectively. Here, the sentence-level and normalized TF-IDF score of a word is considered as the word's weight. Formally, EMD can be defined as:
\begin{equation}
\small
    EMD(\mathcal{H},\mathcal{R})= \frac{\text{min}_{f_{ij}\in\mathcal{F}(\mathcal{H},\mathcal{R})}\sum_{i=1}^{p} \sum_{j=1}^{q} d_{ij}f_{ij}}{\text{min}(m_{\sum},n_{\sum})}\\
    \label{eq:emd}
\end{equation}
where $d_{ij}$ is the distance between the words $r_{i}$ and $h_{j}$ in the space and $\mathcal{F}(\mathcal{H},\mathcal{R})$ is a set of possible flows between the two distributions that the system tries to optimize. In Equation~\ref{eq:emd}, $EMD(\mathcal{H},\mathcal{R})$ denotes the amount of work required to match the hypothesis with the reference. 
The optimization is done following four constraints:
\begin{equation}
\small
\begin{split}
    f_{ij} &\geq 0 \quad\quad i=1,2,...,p\; \text{and}\; j=1,2,..,q,\\
    \sum^{q}_{j=1} f_{ij} &\leq m_{i} \,\quad i=1,2,...,p,\\
    \sum^{p}_{i=1} f_{ij} &\leq n_{j} \,\quad j=1,2,...,q,\\
    \sum^{p}_{i=1}\sum^{q}_{j=1} f_{ij} &= \text{min}(m_{\sum},n_{\sum})\\
\end{split}
\label{eq:constr}
\end{equation}
The first constraint indicates that each flow must be non-negative. The second constraint limits the total weights flowing from $r_{i}$ to less than or equal to $m_{i}$. Similarly, the third constraint restricts the total weights flowing from $h_{j}$ to less than or equal to $n_{j}$. The final constraint indicates that the total flow of weights must be equal to the minimum weight distribution. Figure~\ref{fig:emdintro} depicts the EMD for a given hypothesis-reference pair.

\subsection{Syntactic Similarity and Tree Edit Distance}
In computational linguistics, dependency and constituency trees are used to represent syntactic dependencies between words in a sentence. Unlike the constituency tree, a dependency tree can represent non-adjacent and non-projective dependencies in a sentence, which frequently appear in spoken language and noisy text. That leads us to prefer dependency trees over constituency trees for evaluating NLG output.

Formally, a dependency tree is a set of nodes $\Omega=\{w_{0},w_{1},...,w_{k}\}$ and a set of dependency links $\mathcal{G} =\{g_{0},g_{1},...,g_{k}\}$, where $w_0$ is the imaginary root node and $g_{i}$ is an index into $\Omega$ representing the governor of $w
_{i}$. Every node has exactly one governor except for $w_0$, which has no governor \cite{hall2010corrective}. Syntactic similarity between a pair of dependency trees can be estimated using several methods, such as graph centralities and Euclidean distances~\cite{oya2020syntactic}.
In our work, we exploit the Tree Edit Distance (TED) algorithm~\cite{zhang1989simple} to estimate syntactic similarity between reference and hypothesis.
TED is typically computed on ordered labeled trees and can thus be used to compare dependency trees. The \textit{edit} operations performed during the comparison of parsed dependency trees include \textit{Change}, \textit{Delete}, and \textit{Insert}.
\begin{figure}[htb!]
\centering
\includegraphics[width=6.5cm]{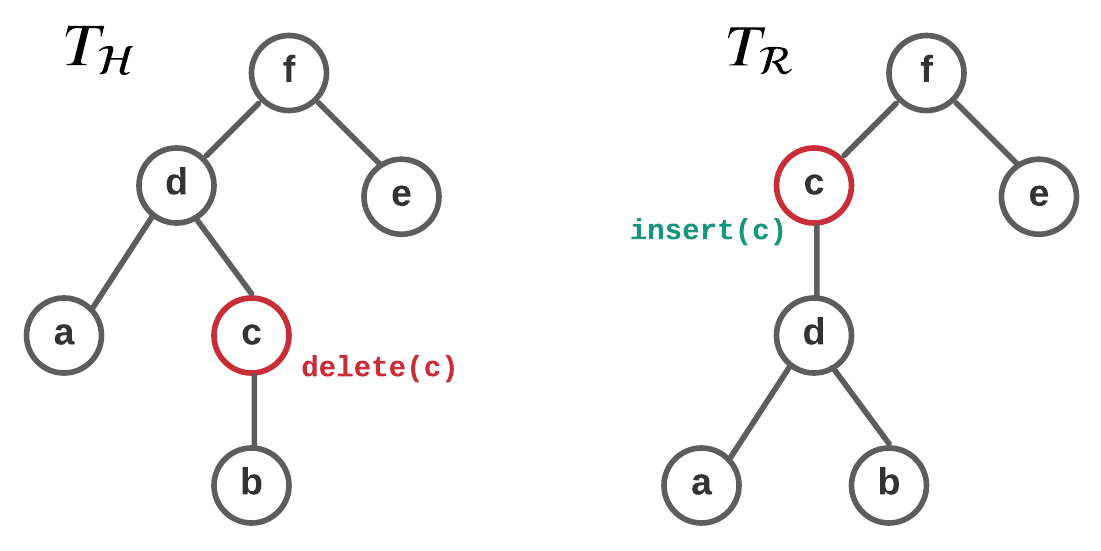}
    \caption{Visualization of the required \textit{edit} operations to transform $T_{\mathcal{H}}$ to $T_{\mathcal{R}}$. The operations corresponds to the following sequence: \textit{delete}(node with label \textit{c}), \textit{insert}(node with label \textit{c}).}
    \label{fig:ted}
\end{figure}

Let us consider $T_{\mathcal{H}}$ and $T_{\mathcal{R}}$ be the parsed dependency trees of the hypothesis and reference, respectively. 
The operations required to transform one tree into another  are visualized in Figure~\ref{fig:ted}.
In TED, an exact match between the nodes of the compared trees is performed to decide if any edit operation is required. In this work, the syntactic difference between hypothesis and reference is determined by the output of TED, which specifies the total number of edit operations.
\section{RoMe}
In RoMe, a neural network determines the final evaluation score given a reference-hypothesis pair. The network is trained to predict the evaluation score based on three features: semantic similarity computed by EMD, enhanced TED, and the grammatical acceptability score. We explain these features in the following subsections.

\subsection{Earth Mover's Distance Based Semantic Similarity}
\label{subsec:emd}
During the computation of EMD, we employ \textit{hard word alignment} and \textit{soft-penalization} techniques to tackle repetitive words and passive forms of a sentence. We compute a distance matrix and a flow matrix as described below and finally obtain EMD utilizing Equation~\ref{eq:emd}.


\textbf{Hard Word Alignment.} 
We first align the word pairs between reference and hypothesis based on their semantic similarities.
The alignment is performed by computing all paired cosine similarities while taking word position information into account, as in~\cite{echizen2019word}. 
In contrast to ~\cite{echizen2019word}, we use contextualized pre-trained word embedding from the language model ALBERT~\cite{albert}. ALBERT uses sentence-order prediction loss, focusing on modeling inter-sentence coherence, which improves multi-sentence encoding tasks.
The word alignment score is computed as follows:
\begingroup\makeatletter\def\f@size{9}\check@mathfonts
\begin{equation}
\mathcal{A}(r_{i},h_{j}) = \frac{\vec{r_{i}}\cdot \vec{h_{j}}}{\|\vec{r_{i}}\| \|\vec{h_{j}}\|} \cdot \frac{|q\left(i+1\right)-p\left(j+1\right)|}{pq} \quad \label{eq:aling}
\end{equation}
\endgroup
where $\vec{r_{i}}$ and $\vec{h_{j}}$ denote the contextualized word embedding of $r_{i}$ and $h_{j}$, respectively. The first part of the right side of the equation computes the cosine similarity between $\vec{r_{i}}$ and $\vec{h_{j}}$, and the second part calculates the relative position information as proposed in~\cite{echizen2019word}.
\begin{figure}[ht!]
\centering
\includegraphics[width=0.8\columnwidth]{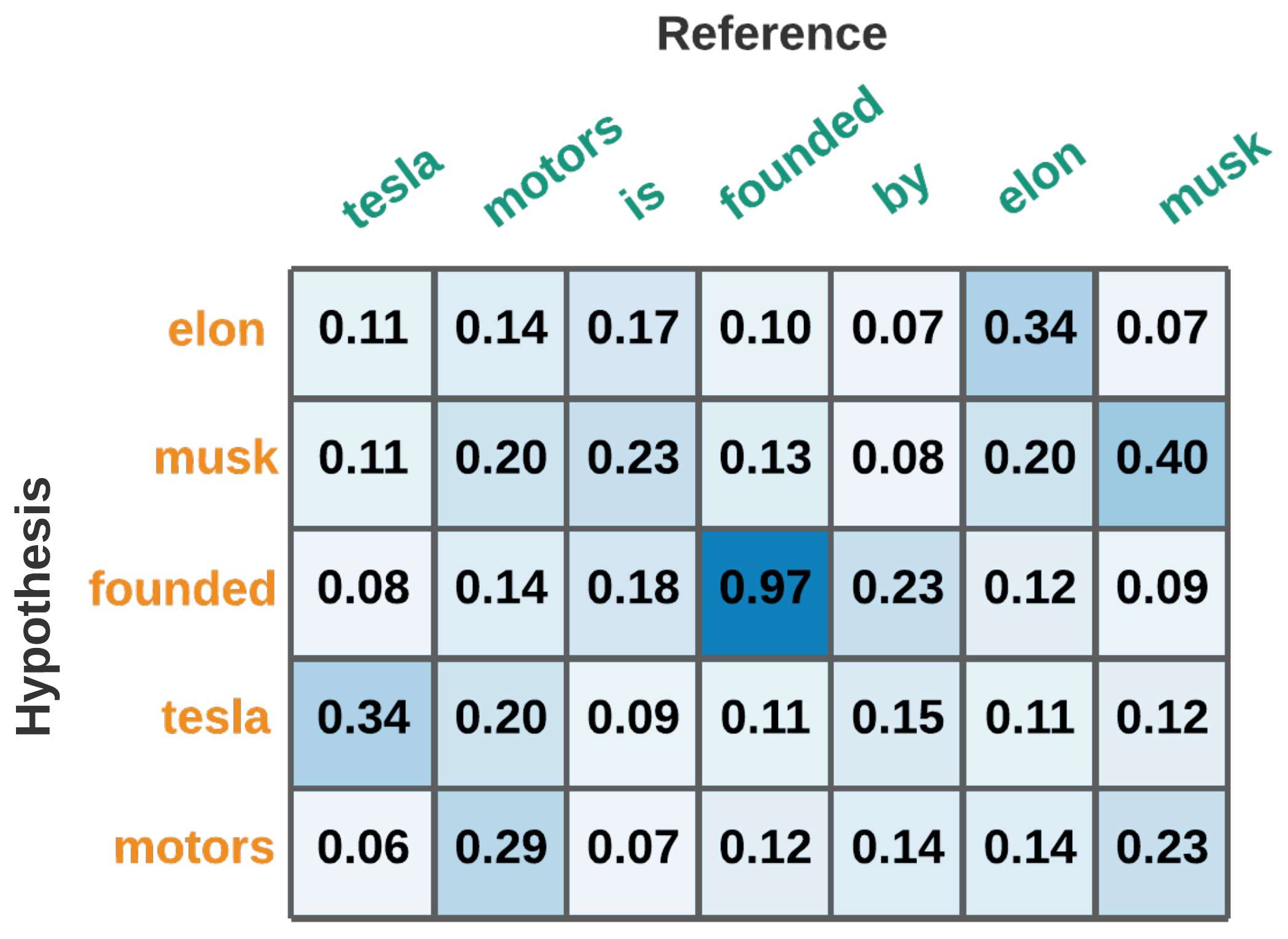}
\caption{An example word alignment matrix for the reference sentence: "\textit{tesla motors is founded by elon musk}" and its passive form: "\textit{elon musk founded tesla motors}" is illustrated here.} 
    \label{fig:align_mat}
\end{figure}

\begin{figure}[htb!]
\centering
\includegraphics[width=6.5cm]{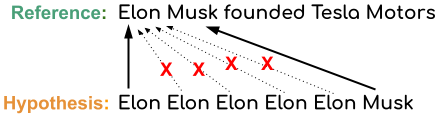}
    \caption{An example hypothesis containing repetitive words.}
    \label{fig:repetive}
\end{figure}

Figure~\ref{fig:align_mat} depicts a matrix of word alignment scores generated on an example pair of sentences. 
This alignment strategy fails to handle repetitive words where a word from the hypothesis may get aligned to several words in the reference (see Figure~\ref{fig:repetive}). To tackle such cases, we restrict the word alignment by imposing a hard constraint.
In the hard constraint, we prevent the words in the hypothesis from getting aligned to multiple words in the reference as illustrated by the dotted arrows in Figure~\ref{fig:repetive}. We denote the resulting set of hard-aligned word pairs as $\mathcal{A}_{hc}$.

\textbf{Transport Distance.} A distance matrix $\mathcal{D}$ is required to compute the final EMD score. For each aligned pair $(r_{i},h_{j})\in \mathcal{A}_{hc}$ where $\frac{\vec{r_{i}}\cdot \vec{h_{j}}}{\|\vec{r_{i}}\| \|\vec{h_{j}}\|}>\delta$, the distance between $r_{i}$ and $h_{j}$ is computed as follows: 
\begin{equation}
\begin{small}
d_{ij} = 1.0-\frac{\vec{r_{i}}\cdot \vec{h_{j}}}{\|\vec{r_{i}}\| \|\vec{h_{j}}\|} \cdot e^{\gamma\cdot\frac{|q\left(i+1\right)-p\left(j+1\right)|}{pq}}
\end{small}
    \label{eq:disteq}
\end{equation}
where $d_{ij}\in\mathcal{D}$ and $\delta$ is a confidence threshold found via hyper-parameter search, $\gamma \in [-1, 0)$ is a soft-penalization constant. For all the non-hard-aligned pairs and aligned pairs with value less than $\delta$, the distance $d_{ij}$ receives a maximum value of 1.0. Intuitively, a lower value of $d_{ij}$ implies that the word needs to travel a shorter distance in the transportation problem of EMD.  In Equation~\ref{eq:disteq}, $e^{\gamma\cdot\frac{|q\left(i+1\right)-p\left(j+1\right)|}{pq}}$ works as a penalty where a higher position difference multiplied with the negative constant $\gamma$ will results in low $d_{ij}$ score. The role of $\gamma$ is explained below.

\textbf{Soft-penalization.} Existing metrics often impose hard penalties for words with different order than the reference sentence~\cite{zhao-etal-2019-moverscore, echizen2019word}. For instance, sentences phrased in the passive form obtain a very low score in those metrics. Addressing this issue, we introduce a soft-penalization constant $\gamma= -\frac{|j-i|}{\text{max}(p,q)}$ in Equation~\ref{eq:disteq} to handle the passive form of a sentence better. Let us consider a reference, "\textit{Shakespeare has written Macbeth}" and the passive form of the sentence as hypothesis, "\textit{The Macbeth is written by Shakespeare}". The word \textit{Shakespeare} appears at the beginning of the reference and at the end of the hypothesis, thus the position difference is larger. 
In such scenario, $\gamma$ imposes a lower penalty as it divides the position difference by the length $max(p,q)$.

Finally, following the optimization constraints of Equation~\ref{eq:constr}, we obtain the transportation flow $\mathcal{F}(\mathcal{H},\mathcal{R})$. 
For the optimized flow $f_{ij}\in \mathcal{F}(\mathcal{H},\mathcal{R})$, the final equation of EMD is as follows:
\begin{equation}
\small
    EMD(\mathcal{H},\mathcal{R}) = \frac{\text{min}_{f_{ij}\in\mathcal{F}(\mathcal{H},\mathcal{R})}\sum_{i=1}^{p} \sum_{j=1}^{q} d_{ij}f_{ij}}{\text{min}(m_{\sum},n_{\sum})}
\end{equation}
The semantic similarity between hypothesis and reference is denoted as $\mathcal{F}_{sem} = 1.0 - EMD$. The normalized value of EMD is used to calculate $\mathcal{F}_{sem}$.

\subsection{Semantically Enhanced TED}
\label{subsec:sted}
To estimate the difference between the syntactic structures of reference and hypothesis, we extend the TED algorithm~\cite{zhang1989simple}. The original TED algorithm performs edit operations based on an exact match between two nodes in the dependency trees of hypothesis and reference. 
In this work, we modify the TED algorithm and compute a word embedding-based cosine similarity to establish the equivalence of two nodes.
Two nodes are considered equal, if the cosine similarity of their embedding representations exceeds the threshold $\theta$. 
This allows the semantically enhanced TED to process synonyms and restricts it from unnecessary editing of similar nodes. We call the resulting algorithm TED-SE. The normalized value of TED-SE is denoted as $\mathcal{F}_{ted}$. 
We compute TED-SE over the lemmatized reference and hypothesis since lemmatized text exhibits improved performance in such use cases~\cite{kutuzov-kuzmenko-2019-lemmatize}. The lemmatizer and dependency parser from Stanza~\cite{qi2020stanza} are utilised to obtain the tree representation of the text.
Further details are provided in Appendix~\ref{app:ted_dt}.

\subsection{Grammatical Acceptability Classification}
\label{subsec:gram}
Linguistic competence assumes that native speakers can judge the grammatical acceptability of a sentence. However, system-generated sentences are not always grammatically correct or acceptable. 
Therefore, we train a binary classifier on the Corpus of Linguistic Acceptability (CoLA)~\cite{warstadt-etal-2019-neural}, predicting  the probability that the hypothesis is grammatically acceptable. CoLA is a collection of sentences
from the linguistics literature with binary expert acceptability labels containing over 10k examples~\cite{warstadt-etal-2019-neural}~\footnote{with 70.5\% examples manually labeled \textit{acceptable}.}. The classifier is based on BERT-large~\cite{devlin-etal-2019-bert} and trained to optimize binary cross-entropy loss. A text sequence is fed as input and as output, the classifier produces the class membership probability (grammatically acceptable, grammatically unacceptable). 
The model achieves an accuracy of 80.6\% 
on the out-of-domain CoLA test set ~\cite[p.~8]{warstadt-etal-2019-neural}. 
We denote the score from the classifier as the feature $\mathcal{F}_{g}$, which is used to train a neural network (see~\textsection{\ref{subsec:net}}). 

\begin{table*}[!htb]
\begin{minipage}{.67\linewidth}
    \medskip
\begin{adjustbox}{width=\textwidth, center}
\begin{tabular}{l|l|ccc|ccc}
\toprule
\multirow{2}{*}{\textbf{Settings}}& \multicolumn{1}{c|}{\multirow{2}{*}{\textbf{Metrics}}} & \multicolumn{3}{c|}{\textbf{BAGEL}}         & \multicolumn{3}{c}{\textbf{SFHOTEL}}       \\
                  & \multicolumn{1}{c|}{}                                  & \textbf{Info} & \textbf{Nat} & \textbf{Qual} & \textbf{Info} & \textbf{Nat} & \textbf{Qual} \\ \midrule
                  & BLEU-1                                        & 0.225        & 0.141        & 0.113         & 0.107        & 0.175        & 0.069         \\
                  & BLEU-2                                        & 0.211        & 0.152        & 0.115         & 0.097        & 0.174        & 0.071         \\
                  & METEOR                                        & 0.251        & 0.127        & 0.116         & 0.163        & 0.193        & 0.118         \\
                  & BERTScore                                  & 0.267        & 0.210        & 0.178         & 0.163        & 0.193        & 0.118         \\ \cline{2-8}
                  & SMD+W2V                                       & 0.024        & 0.074        & 0.078         & 0.022        & 0.025        & 0.011         \\
Baselines         & SMD+ELMO+PMEANS                               & 0.251        & 0.171        & 0.147         & 0.130        & 0.176        & 0.096         \\
                  & SMD+BERT+MNLI+PMAENS                          & 0.280        & 0.149        & 0.120         & 0.205        & 0.239        & 0.147         \\ \cline{2-8}
                  & WMD-1+ELMO+PMEANS                             & 0.261        & 0.163        & 0.148         & 0.147        & 0.215        & 0.136         \\
                  & WMD-1+BERT+PMEANS                             & \cellcolor{blue!20}0.298        & 0.212        & 0.163         & 0.203        & 0.261        & 0.182         \\
                  & WMD-1+BERT+MNLI+PMEANS                        & 0.285        & 0.195        & 0.158         & 0.207        & 0.270        & 0.183         \\ \midrule
                  & RoMe (Fasttext)                               &      0.112        &       0.163       &       0.132        &      0.172        &       0.190       &   0.231            \\
RoMe              & RoMe (BERT)                                    &      0.160        &       0.251       &     0.202          &     0.212         &      0.283        &             0.300  \\
                  & RoMe (ALBERT-base)                            &       0.162       &     0.259         &       0.222       &          0.231    &      0.295        &  0.315             \\ \cline{2-8}
                  & \textbf{RoMe (ALBERT-large)}                          &      0.170        &      \cellcolor{blue!20}0.274       &   \cellcolor{blue!20}0.241             & \cellcolor{blue!20}0.244             &     \cellcolor{blue!20}0.320         &         \cellcolor{blue!20}0.327     \\ \bottomrule
\end{tabular}
\end{adjustbox}
\caption{Spearman correlation ($\rho$) scores computed from the metric scores with respect to the human evaluation scores on BAGEL and SFHOTEL. Baseline model's results are reported form~\cite{zhao-etal-2019-moverscore}. Here, \textbf{Info}, \textbf{Nat} and \textbf{Qual} refer to \textit{informativeness}, \textit{naturalness}, and \textit{quality}, respectively.}
\label{tab:bagelsfhotel}
\end{minipage}\hfill
\begin{minipage}{.3\linewidth}
\begin{figure}[H]
\includegraphics[width=\columnwidth]{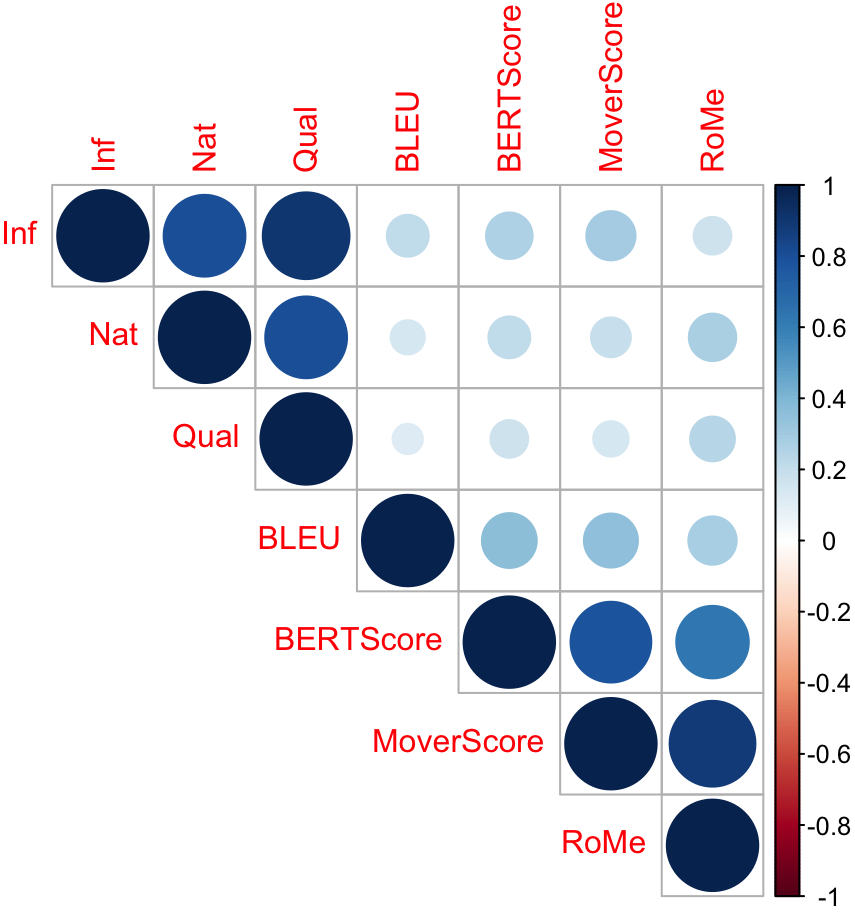}
\caption{Correlation between the explored metrics.}
\label{fig:corrplot}
\end{figure}
\end{minipage}
\end{table*}

\subsection{Final Scorer Network}
\label{subsec:net}
A feed-forward neural network takes the previously computed features as input and learns a function $f(\mathcal{F}_{sem}; \mathcal{F}_{ted}; \mathcal{F}_{g} )$ in the final step, yielding a final output score in the $[0, 1]$ interval. The output score is regarded as the overall quality of the hypothesis. Following a self-supervised paradigm, the network is trained on artificially generated training samples from the KELM dataset~\cite{agarwal-etal-2021-knowledge}. KELM contains knowledge-grounded natural sentences. 
We randomly choose 2,500 sentence pairs from the KELM dataset and generate 2,500 more negative samples by randomly augmenting the sentences using TextAttack~\cite{morris2020textattack} and TextFooler~\cite{jin2020bert}. Following a similar approach, we additionally generate 1,000 test sentence pairs from the KELM dataset. Overall, we then have 5,000 training and 1,000 test examples. 
The network is a simple, two-layered feed-forward network optimized with stochastic gradient descent using a learning rate of 1e-4.

\begin{table*}[]
\begin{adjustbox}{width=0.75\textwidth,center}
\begin{tabular}{l|l|cccc}
\toprule
       & \textbf{Text}                                                                            & \textbf{EMD}        & \textbf{TED-SE}                            & \textbf{Grammar}        & \textbf{RoMe}              \\ \midrule
\multicolumn{1}{l|}{$\mathcal{R}$} & Munich is located at the southern part of Germany.   & \multicolumn{1}{c}{\multirow{2}{*}{0.83}} & \multirow{2}{*}{1.0} & \multirow{2}{*}{0.94} & \multirow{2}{*}{0.80} \\
\multicolumn{1}{l|}{$\mathcal{H}$} & Munich is situated in the south of Germany. & \multicolumn{1}{c}{}                   &                   &                   \\ \hline
\multicolumn{1}{l|}{$\mathcal{R}$} & Tesla motors is founded by Elon Musk.      & \multirow{2}{*}{0.70}                       & \multirow{2}{*}{0.85} & \multirow{2}{*}{0.96} & \multirow{2}{*}{0.69} \\
\multicolumn{1}{l|}{$\mathcal{H}$} & Elon Musk has founded Tesla Motors.   &                                         &                   &                   \\ \hline
\multicolumn{1}{l|}{$\mathcal{R}$} & Elon musk has founded tesla motors.  & \multirow{2}{*}{0.01}                       & \multirow{2}{*}{0.50} & \multirow{2}{*}{0.17} & \multirow{2}{*}{0.11}\\
\multicolumn{1}{l|}{$\mathcal{H}$} & Elon elon elon elon elon founded tesla tesla tesla. &                                         &                   &          \\\bottomrule       
\end{tabular}
\end{adjustbox}
  \caption{Component-wise qualitative analysis.}
  \label{tab:qualcomp}
  \vspace{-0.4cm}
\end{table*}
\begin{table*}[]
\begin{adjustbox}{width=\textwidth,center}
\begin{tabular}{l|l|cccc}
\toprule
       & \textbf{Text}                                                                            & \textbf{BLEU}        & \textbf{BERTScore}                            & \textbf{MoverScore}        & \textbf{RoMe}              \\ \midrule
\multicolumn{1}{l|}{$\mathcal{R}$} & James Craig Watson, who died from peritonitis, discovered 101 Helena.   & \multicolumn{1}{c}{\multirow{2}{*}{0.0}} & \multirow{2}{*}{0.81} & \multirow{2}{*}{0.54} & \multirow{2}{*}{0.15} \\
\multicolumn{1}{l|}{$\mathcal{H}$} & The Polish Academy of Science is regionserved.  & \multicolumn{1}{c}{}                   &                   &                   \\ \hline
\multicolumn{1}{l|}{$\mathcal{R}$} & 1001 gaussia was formerly known as 1923 oaa907 xc.     & \multirow{2}{*}{0.0}                       & \multirow{2}{*}{0.79} & \multirow{2}{*}{0.51} & \multirow{2}{*}{0.13} \\
\multicolumn{1}{l|}{$\mathcal{H}$} & The former name for the former name for 11 gunger is 1923. One of the former name is 1923.   &                                         &                   &             \\\bottomrule       
\end{tabular}
\end{adjustbox}
  \caption{Qualitative analysis.}
  \label{tab:qualana}
\end{table*}

\begin{table*}[!htb]
\begin{minipage}{.68\linewidth}
    \medskip
\begin{adjustbox}{width=0.9\textwidth,center}
  \begin{tabular}{l|c|c|c|c|c|c|c}
  \toprule
      \multicolumn{1}{l|}{\textbf{Dialogue dataset}} &
      \multicolumn{1}{c|}{\textbf{Models}} &
      \multicolumn{1}{c|}{\textbf{SentBLEU}} &
      \multicolumn{1}{c|}{\textbf{METEOR}} &
      \multicolumn{1}{c|}{\textbf{BERTScore}} &
      \multicolumn{1}{c|}{\textbf{MoverScore}} & \multicolumn{1}{c|}{\textbf{RoMe}}\\
    \hline
   & Mem2Seq& 0.07 & 0.35 & 0.40 & 0.49 & \cellcolor{blue!20}0.51\\
        \textbf{In-car dialogue} & GLMP & 0.04& 0.29 &\cellcolor{blue!20} 0.32 & 0.31 &\cellcolor{blue!20}0.32 \\
            \textbf{} & DialoGPT  & 0.17 & 0.60 & 0.62 & 0.73 & \cellcolor{blue!20}0.78\\
    \hline
        \textbf{ } & Mem2Seq  & 0.03 & 0.08 & 0.08 & \cellcolor{blue!20}0.11 & \cellcolor{blue!20}0.11\\
        \textbf{Soccer dialogue} & GLMP & 0.02 & 0.08 & 0.03 & 0.12 &\cellcolor{blue!20} 0.14\\
        \textbf{} & DialoGPT   & 0.04 & 0.26 & 0.31 &0.39 & \cellcolor{blue!20}0.43\\
        \bottomrule
  \end{tabular}
  
  \end{adjustbox}
    \caption{Metrics Spearman's correlation coefficient ($\rho$) with human judgment on dialogue datasets.\label{tab:dialo}}
\end{minipage}
\begin{minipage}{.37\linewidth}
\begin{adjustbox}{width=0.79\columnwidth,center}
\begin{tabular}{l|c}
\toprule
\textbf{Approaches} & \textbf{Correlation ($\rho$)} \\ \hline
RoMe with $\text{EMD}_{std}$ & 64.8 \\
\quad\quad \;\;\;+ $\text{EMD}_{align}$ & 66.0 \\
\quad\quad \;\;\;+ $\text{EMD}_{soft}$ & 66.9 \\
\quad\quad \;\;\;+ TED-SE &  69.1 \\
\quad\quad \;\;\;+ Grammar & 70.1\\ \bottomrule
\end{tabular}
\end{adjustbox}
\caption{Ablation Study.\label{tab:ablation}}
\end{minipage}
\end{table*}

\section{Experiments and Analysis}
\subsection{Data}
To assess RoMe's overall performance, first, we benchmark on two language generation datasets, BAGEL~\cite{bagel} and SFHOTEL~\cite{wen-etal-2015-semantically}, containing 404 and 796 data points, respectively. Each data point contains a meaning representation (MR) and a system generated output. Human evaluation scores of these datasets are obtained from~\cite{novikova-etal-2017-need}. 
Furthermore, we evaluate dialogue system's outputs on Stanford in-car dialogues~\cite{eric2017key} containing 2,510 data points and the soccer dialogue dataset~\cite{chaudhuri2019using} with 2,990 data points. 
Each data point of these datasets includes a user query, a reference response, and a system response as a hypothesis. Three different system outputs are evaluated for each dialogue dataset. 
We use the human annotated data provided by~\cite{chaudhuri2021grounding}.  Moreover, we evaluate the metrics on the system generated outputs from the WebNLG 2017 challenge~\cite{shimorina-2020-webnlg-human-eval}.

Finally, to conduct robustness analysis, we randomly sample data points from KELM~\cite{agarwal-etal-2021-knowledge} and perturb them with adversarial text transformation techniques. 
Three annotators participated in the data annotation process (two of them are from a Computer Science and one from a non-Computer Science background), where they annotated the perturbed data. 
We provided the annotators with an annotation tool which displays the reference sentence and the system output for each data point. The annotators were asked to choose a value from a range of [1,3], for each of the categories: \textit{Fluency}, \textit{Semantic Correctness}, and \textit{Grammatical correctness}. In this case, the values stand for 1: \textit{poor}, 2: \textit{average}, and 3: \textit{good}. The overall inter-annotator agreement score, $\kappa$ is 0.78. The annotation tool and its interface are discussed in detail in Appendix~\ref{app:anno}.

\subsection{Hyper-parameter Settings}
We use $\delta=0.60$ and $\theta=0.65$ in \textsection{\ref{subsec:emd}}. Best values are found by a hyper-parameter search from a range of [0,1.0] with an interval of 0.1. RoMe obtained the best result by utilizing ALBERT-large~\cite{albert} model with 18M parameters and 24 layers. Furthermore, we use the English word embedding of dimension 300 to obtain results from Fasttext~\cite{bojanowski2017enriching} throughout the paper. As the grammatical acceptability classifier, we train a BERT-base model with 110M parameters and 12 layers. The hidden layer size is 768 with a hidden layer dropout of 0.1. A layer norm epsilon of 1e-12 was used for layer normalization. GELU~\cite{hendrycks2016gaussian} was used as the activation function. We use a single GPU with 12GBs of memory for all the evaluations.


\subsection{Baselines}
We select both the word-overlap and embedding-based metrics as strong baselines. For the experiment and robustness analysis we choose BLEU~\cite{papineni2002bleu}, METEOR~\cite{banerjee2005meteor}, BERTScore~\cite{zhang-2020BERTScore} and MoverScore~\cite{zhao-etal-2019-moverscore}. We evaluate the metrics on the sentence level to make a fair comparison.
\begin{table*}[ht!]
\begin{adjustbox}{width=0.9\textwidth,center}
  \begin{tabular}{l|c|c|c|c|c|c|c|c|c|c|c|c|c|c|c}
    \toprule
    \multicolumn{1}{l|}{\textbf{Metrics}} &
      \multicolumn{3}{c|}{\textbf{BLEU}} &
      \multicolumn{3}{c|}{\textbf{METEOR}} &
      \multicolumn{3}{c|}{\textbf{BERTScore}} &
      \multicolumn{3}{c}{\textbf{MoverScore}} &
      \multicolumn{3}{|c}{\textbf{RoMe}}\\
      \hline
   \textbf{Systems} & $\rho$ & \textit{r} & $\tau$ & $\rho$ & \textit{r} & $\tau$ & $\rho$ & \textit{r} & $\tau$ & $\rho$ & \textit{r} & $\tau$& $\rho$ & \textit{r} & $\tau$\\
    \hline
    ADAPT & 0.38 & 0.39 & 0.27 & 0.57 & 0.58 & 0.41 & 0.61 & 0.72 & 0.50 & 0.68 & \cellcolor{blue!20}0.73 & 0.49 &\cellcolor{blue!20} 0.72 & 0.70 & \cellcolor{blue!20}0.51\\
    \hline
    Baseline & 0.35 & 0.42 & 0.26 & 0.49 & 0.49 & 0.33  & 0.49 & 0.50 & 0.35 & \cellcolor{blue!20}0.59 & \cellcolor{blue!20}0.61 &\cellcolor{blue!20} 0.43 & 0.53 & 0.53 & 0.37\\
        \hline
        melbourne & 0.32 & 0.31 & 0.21 & 0.35 & 0.35 & 0.24 &  0.33 & 0.33  & 0.26 & 0.40 & 0.39 & 0.28 &\cellcolor{blue!20} 0.44 &\cellcolor{blue!20}0.50 & \cellcolor{blue!20}0.35\\
    \hline
        Pkuwriter & 0.37 & 0.38 & 0.28 & 0.47 & 0.47 & 0.31 &  0.48 & 0.53 & 0.38 & 0.57 & \cellcolor{blue!20}0.56 & \cellcolor{blue!20}0.39 & \cellcolor{blue!20}0.58 & \cellcolor{blue!20}0.56 & \cellcolor{blue!20}0.39 \\
    \hline
    tilburg-nmt & 0.25 & 0.20 & 0.13 & 0.26 & 0.26 & 0.18 & 0.38 & 0.39 & 0.30 & 0.49 & 0.50 & 0.36 &\cellcolor{blue!20} 0.64 & \cellcolor{blue!20}0.68 & \cellcolor{blue!20}0.50\\
    \hline
    tilburg-pipe & 0.38 & 0.41 & 0.30 & 0.52 & 0.43 & 0.30 & 0.53 & 0.48 & 0.33 &\cellcolor{blue!20} 0.62 & \cellcolor{blue!20}0.50 & \cellcolor{blue!20}0.35 & 0.38 & 0.42 & 0.27\\
        \hline
    tilburg-smt & 0.25 & 0.20 & 0.13 & 0.21 & 0.19 & 0.13 & 0.33 & 0.30 & 0.25 & 0.40 & 0.38 & 0.27 & \cellcolor{blue!20}0.50 & \cellcolor{blue!20}0.51 & \cellcolor{blue!20}0.36\\
        \hline
    upf-forge & 0.14 & 0.13 & 0.08 & 0.13 & 0.11 & 0.08 & 0.26 & 0.25 & 0.19 & 0.27 & 0.27 & 0.18 & \cellcolor{blue!20}0.42 & \cellcolor{blue!20}0.42 & \cellcolor{blue!20}0.30\\
        \hline
    vietnam & 0.73 & 0.80 & 0.62 & 0.87 & 0.90 \cellcolor{blue!20}& 0.72 & 0.81 & 0.76 & 0.70 & \cellcolor{blue!20}0.90 & 0.78 & 0.73 & 0.84 & 0.89 & \cellcolor{blue!20}0.83\\
    \bottomrule
  \end{tabular}
  \end{adjustbox}
  \caption{Metrics correlation with human judgment on system outputs from the WebNLG 2017 challenge. Here, \textit{r}: Pearson correlation co-efficient, $\rho$: Spearman's correlation co-efficient, $\tau$: Kendall's Tau.}
    \label{tab:nlg17_comparison}
    \vspace{-0.3cm}
\end{table*}
\begin{table*}[]
\begin{adjustbox}{width=\textwidth,center}
  \begin{tabular}{l|c|c|c|c|c|c|c|c|c|c|c|c|c|c|c}
    \toprule
    \multicolumn{1}{l|}{\textbf{Metrics}} &
      \multicolumn{3}{c|}{\textbf{BLEU}} &
      \multicolumn{3}{c|}{\textbf{METEOR}} &
      \multicolumn{3}{c|}{\textbf{BERTScore}} &
      \multicolumn{3}{c}{\textbf{MoverScore}} &
      \multicolumn{3}{|c}{\textbf{RoMe}}\\
      \hline
   \textbf{Perturbation methods} & \textit{f} & \textit{s} & \textit{g} & \textit{f} & \textit{s} & \textit{g} & \textit{f} & \textit{s} & \textit{g} & \textit{f} & \textit{s} & \textit{g}& \textit{f} & \textit{s} & \textit{g}\\
    \hline
    Entity replacement & 0.06 & 0.04 & 0.06 & 0.09 & 0.09 & 0.08 & 0.11 & 0.07 & 0.09 &\cellcolor{blue!20}0.16 & 0.13 & 0.11 & \cellcolor{blue!20}0.16 & \cellcolor{blue!20}0.19 & \cellcolor{blue!20}0.14\\
    \hline
    Adjective replacement & 0.07& 0.06 &0.07 & 0.09 & 0.13 & 0.11 & 0.11 & 0.11 & 0.13 & 0.13 & 0.17 & 0.16 &\cellcolor{blue!20} 0.18 & \cellcolor{blue!20}0.23 & \cellcolor{blue!20}0.18\\
        \hline
        Random word replacement & 0.05 & 0.06 & 0.03 & 0.06 & 0.06 & 0.05 & 0.11 & 0.10 & 0.08 & 0.11 & 0.13 & 0.09 & \cellcolor{blue!20}0.15 & \cellcolor{blue!20}0.15 & \cellcolor{blue!20}0.23\\
    \hline
        Text transformation & 0.03 & 0.01 & 0.03 &  0.08 & 0.09 & 0.07 & 0.13 & 0.15 & 0.15 & 0.15 & 0.18 & 0.19 & \cellcolor{blue!20}0.18 & \cellcolor{blue!20}0.19 & \cellcolor{blue!20}0.21\\
    \hline
    Passive form & 0.02 & 0.01 & 0.04 & 0.08 & 0.10 & 0.08 &  0.19 & 0.24 & 0.21 & 0.23 & 0.24 & 0.22 & \cellcolor{blue!20}0.25 & \cellcolor{blue!20}0.28 & \cellcolor{blue!20}0.28\\
    \bottomrule
  \end{tabular}
  \end{adjustbox}
  \caption{Metrics Spearman correlation score against human judgment on perturbed texts. Here, \textit{f}: fluency, \textit{s}: semantic similarity, \textit{g}: grammatical correctness.}
    \label{tab:robann}
    \vspace{-0.3cm}
\end{table*}

\subsection{Results}
Table~\ref{tab:bagelsfhotel} shows the performance of different metrics on data to language generation datasets (BAGEL and SFHOTEL). In both the BAGEL and SFHOTEL, a meaning representation (MR), for instance \textit{inform(name='hotel drisco',price\_range='pricey')} is given as a reference sentence, where the system output is: \textit{the hotel drisco is a pricey hotel}, in this case. Although, RoMe outperformed the baseline metrics in evaluating the \textit{informativeness}, \textit{naturalness} and \textit{quality} score, the correlation scores remain low with regard to human judgment. This is because the MR, which is not a natural sentence, is the reference statement in this scenario. For all the experiments, we take the normalized human judgement scores. We firstly evaluate our model using Fasttext~\cite{bojanowski2017enriching} word embedding. We notice a significant improvement in results when we replace the Fasttext embedding with contextualized word embedding obtained from BERT~\cite{devlin-etal-2019-bert}. Furthermore, we experiment with multiple language models and finally, we reach to our best performing model with ALBERT-large~\cite{albert}. In all the experiments, we report the results of RoMe, using  ALBERT-large~\cite{albert}. In Table~\ref{tab:bagelsfhotel}, WMD and SDM refer to word mover distance and sentence mover distance, respectively, used in MoverScore. We report the results of WDM and SMD from~\cite{zhao-etal-2019-moverscore}.

Table~\ref{tab:dialo} demonstrates the evaluation results on dialogue datasets. We evaluated the system-generated dialogues from three dialogue system models: Mem2Seq~\cite{madotto2018mem2seq}, GLMP~\cite{wu2019global}, and DialoGPT~\cite{zhang-etal-2020-dialogpt}. In case of in-car dataset, all the non-word-overlap metric achieved a better correlation score than the word-overlap based metrics. This is because generated responses in dialogue systems are assessed based on the overall semantic meaning and correctness of the information. 
Overall, RoMe achieves stronger correlation scores on both in-car and soccer dialogue datasets in evaluating several dialogue system outputs.

Finally, we investigate the outputs of nine distinct systems that competed in the WebNLG 2017 competition and report the correlation scores in Table~\ref{tab:nlg17_comparison}. Although RoMe achieves the best correlation in most of the cases, we notice a comparable and in some cases better results achieved by the MoverScore~\cite{zhao-etal-2019-moverscore}. 
A correlation graph is plotted in Figure~\ref{fig:corrplot} to investigate the metrics' performance correlations further. The graph is constructed from RoMe and baseline metrics' scores on the BAGEL dataset. As observed from the correlation graph, we can infer that our proposed metric, RoMe correlates highly with the MoverScore. However, since RoMe handles both the syntactic and semantic properties of the text it achieved better results in all the datasets across different NLG tasks.  

\subsection{Ablation Study}
We conduct an ablation study to investigate the impact of the RoMe's components on its overall performance. Table~\ref{tab:ablation} exhibits the incremental improvement in Spearman's correlation coefficient, that each of the components brings to the metric. We randomly choose 100 system-generated dialogue utterances from the dialogue datasets, since they frequently contain sentences in passive form and repetitive words. The correlation of standard EMD with the human judgement is denoted as "RoMe score with $\text{EMD}_{std}$". Inclusion of semantic word alignment ($\text{EMD}_{align}$) and soft-penalization ($\text{EMD}_{soft}$) further improved the correlation score. The classifier was not used until this point in the ablation since there was just one score. 
Moreover, the correlation score improved significantly when the semantically enhanced TED and grammatical acceptability were introduced as features in addition to the EMD score to a neural classifier. We hypothesize that the inclusion of language features related to grammar and syntactic similarity helped the neural network achieve better performance.

\subsection{Qualitative Analysis}
RoMe is developed in a modular fashion, so it may be used to generate scores for semantic similarity, syntactic similarity, and grammatical acceptability separately. Table~\ref{tab:qualcomp} shows the component-wise score and the final score of RoMe on three example data points. In the first example, RoMe demonstrates its ability of capturing similar sentences by obtaining high score. The scores from several components in the second example demonstrate RoMe's ability to handle passive form. The final example in Table~\ref{tab:qualcomp} demonstrates that RoMe penalizes sentence with repetitive word.

Table~\ref{tab:qualana} shows the performance of the three baselines and RoMe in handling erroneous cases. Although the first example contains a completely different hypothesis and the second case with repetitive hypothesis both BERTScore and MoverScore exhibit high score. On the contrary, BLEU score is unable to handle such scenarios. However, by obtaining low scores, RoMe demonstrates its ability to understand such cases better.

\subsection{Robustness Analysis}
In this section, we design five test cases to stress the models' capabilities. For the analysis purpose, we randomly sample data points from KELM~\cite{agarwal-etal-2021-knowledge} (cases 1, 2, and 4) and BAGEL~\cite{bagel} (cases 3 and 5). The annotators annotate the sampled data points on the following criteria: \textit{fluency}, \textit{semantic correctness}, \textit{grammatical correctness}.

\paragraph{Case 1: Entity replacement.}
We perform invariance test (INV) from~\cite{ribeiro-etal-2020-beyond} to check the metrics' NER capability in assessing the text quality. In this approach, we replace the entities present in the text partially or fully with other entities in the dataset. For instance, "\textit{The population of Germany}" gets transformed to "\textit{The population of England}".
\paragraph{Case 2: Adjective replacement.}
Similar to the entity replacement, in this case we choose 100 data points from KELM that contain adjective in them. Then we replace the adjectives with a synonym and an antonym word to generate two sentences from a single data point. For instance, the adjective \textit{different} is replaced with \textit{unlike} and \textit{same}. At the end of this process, we obtain 200 data points.
\paragraph{Case 3: Random word replacement.}
The words in different positions in the text are replaced by a generic token \textit{AAA} following the adversarial text attack method from~\cite{morris2020textattack}, in this case. For instance, the sentence, "\textit{x is a cheap restaurant near y}" is transformed into "\textit{x is a cheap restaurant AAA AAA}". We select the greedy search method with the constraints on stop-words modification from the TextAttack tool. This approach generates repetitive words when two consecutive words are replaced.
\paragraph{Case 4: Text transformation.}
We leverage TextFooler~\cite{jin2020bert} to replace two words in the texts by similar words, keeping the semantic meaning and grammar preserved. 
\paragraph{Case 5: Passive forms.}
In this case, we randomly choose 200 data points from the KELM~\cite{agarwal-etal-2021-knowledge} dataset where the system generated responses are in passive form.

From the results of robustness analysis in Table~\ref{tab:robann}, it is evident that almost all the metrics obtain very low correlation scores with respect to human judgment. Word-overlap based metrics such as BLEU and METEOR mostly suffer from it. Although RoMe achieves higher correlation scores in most of the cases, there are still scope for improvement in handling the fluency of the text better. Text perturbation techniques used to design the test cases often generate disfluent texts. In some cases, the texts' entities or subjects get replaced by words from out of the domain. From our observation, we hypothesize that handling keywords such as entities may lead to a better correlation score.

\section{Related Work}
A potentially good evaluation metric is one that correlates highly with human judgment.
Among the unsupervised approaches, BLEU~\cite{papineni2002bleu}, METEOR~\cite{banerjee2005meteor} and ROUGE~\cite{lin-2004-rouge} are the most popular evaluation metrics traditionally used for evaluating NLG systems.
Although these metrics perform well in evaluating machine translation (MT) and summarization tasks,~\cite{liu-etal-2016-evaluate} shows that none of the word overlap based metrics is close to human level performance in dialogue system evaluation scenarios.
In a different line of work, word embedding based metrics are introduced for evaluating NLG systems~\cite{mikolov2013distributed,matsuo2017word}. 
Several unsupervised automated metrics were proposed that leverage EMD; one of them is word mover's distance (WMD)~\cite{kusner2015word}. 
Later, ~\cite{matsuo2017word} proposed an evaluation metric, incorporating WMD and word-embedding, where they used word alignment between the reference and hypothesis to handle the word-order problem. 
Recently,~\cite{echizen2019word} introduced an EMD-based metric WE\_WPI that utilizes the word-position information to tackle the differences in surface syntax in reference and hypothesis. 

Several supervised metrics were also proposed for evaluating NLG. ADEM~\cite{lowe-etal-2017-towards} uses a RNN-based network to predict the human evaluation scores. 
With the recent development of language model-based pre-trained models~\cite{zhang-2020BERTScore} proposed BERTScore, which uses a pre-trained BERT model for evaluating various NLG tasks such as machine translation and image captions. Recently,~\cite{zhao-etal-2019-moverscore} proposed MoverScore, which utilizes contextualized embedding to compute the mover's score on word and sentence level. A notable difference between MoverScore and BERTScore is that the latter relies on hard alignment compared to soft alignments in the former. 
Unlike the previous methods, RoMe focuses on handling the sentence's word repetition and passive form when computing the EMD score. Furthermore, RoMe trains a classifier by considering the sentence's semantic, syntactic, and grammatical acceptability features to generate the final evaluation score.
\section{Conclusion}
We have presented RoMe, an automatic and robust evaluation metric for evaluating a variety of NLG tasks.
The key contributions of RoMe include 1) \textbf{EMD-based semantic similarity}, where \textit{hard word alignment} and \textit{soft-penalization} techniques are employed into the EMD for tackling repetitive words and passive form of the sentence,
2) \textbf{semantically enhanced TED} that computes the syntactic similarity based on the node-similarity of the parsed dependency trees, 
3) \textbf{grammatical acceptability classifier}, which evaluates the text's grammatical quality, and 4) \textbf{robustness analysis}, which assesses the metric's capability of handling various form of the text. 
Both quantitative and qualitative analyses exhibit that RoMe highly correlates with human judgment. We intend to extend RoMe by including more languages in the future.


\section*{Acknowledgements}
We acknowledge the support of the following projects: SPEAKER (BMWi FKZ 01MK20011A), JOSEPH (Fraunhofer Zukunftsstiftung), OpenGPT-X (BMWK FKZ 68GX21007A), the excellence clusters ML2R (BmBF FKZ 01 15 18038 A/B/C), ScaDS.AI (IS18026A-F) and TAILOR (EU GA 952215).
\bibliography{acl_latex}
\bibliographystyle{acl_natbib}

\clearpage
\appendix

\section{Appendix}
\label{sec:appendix}


\subsection{Dependency Tree Representation for Tree Edit Distance Calculation}
\label{app:ted_dt}
\begin{figure*}[]
\includegraphics[width=\textwidth]{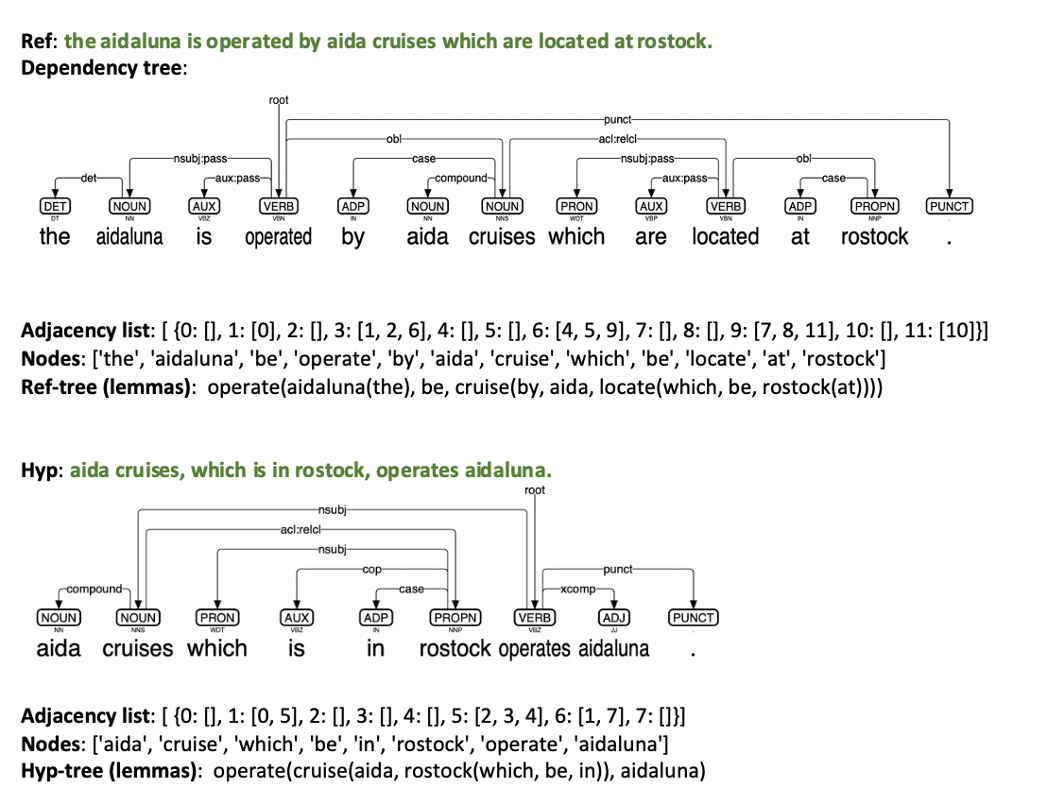}
\caption{Dependency trees of reference and hypothesis, pre-processed for the TED-SE calculation.}
    \label{fig:ted_dt}
\end{figure*}

This section describes the process of parsing a dependency tree from a sentence, followed by converting the dependency tree to the adjacency list for computing TED-SE. Let us consider a reference statement "\textit{the aidaluna is operated by aida cruises which are located at rostock.}" and a hypothesis, "\textit{aida cruises, which is in rostock, operates aidaluna.}".
First, a dependency tree is parsed utilizing the Stanza dependency parser~\cite{qi2020stanza} and then converted to an adjacency list. The adjacency list contains a key-value pair oriented data structure where each key corresponds to a node's index in the tree, and the value is a list of edges on which the head node is incident. Figure~\ref{fig:ted_dt} demonstrates the dependency trees and their corresponding adjacency lists for the given reference and hypothesis. List of nodes and adjacency lists are then fed into the TED-SE algorithm to calculate semantically enhanced tree edit distance as described in~\textsection{\ref{subsec:sted}}.    


\subsection{Annotation Tool}
\label{app:anno}
\begin{figure*}[]
\centering
\includegraphics[width=12cm]{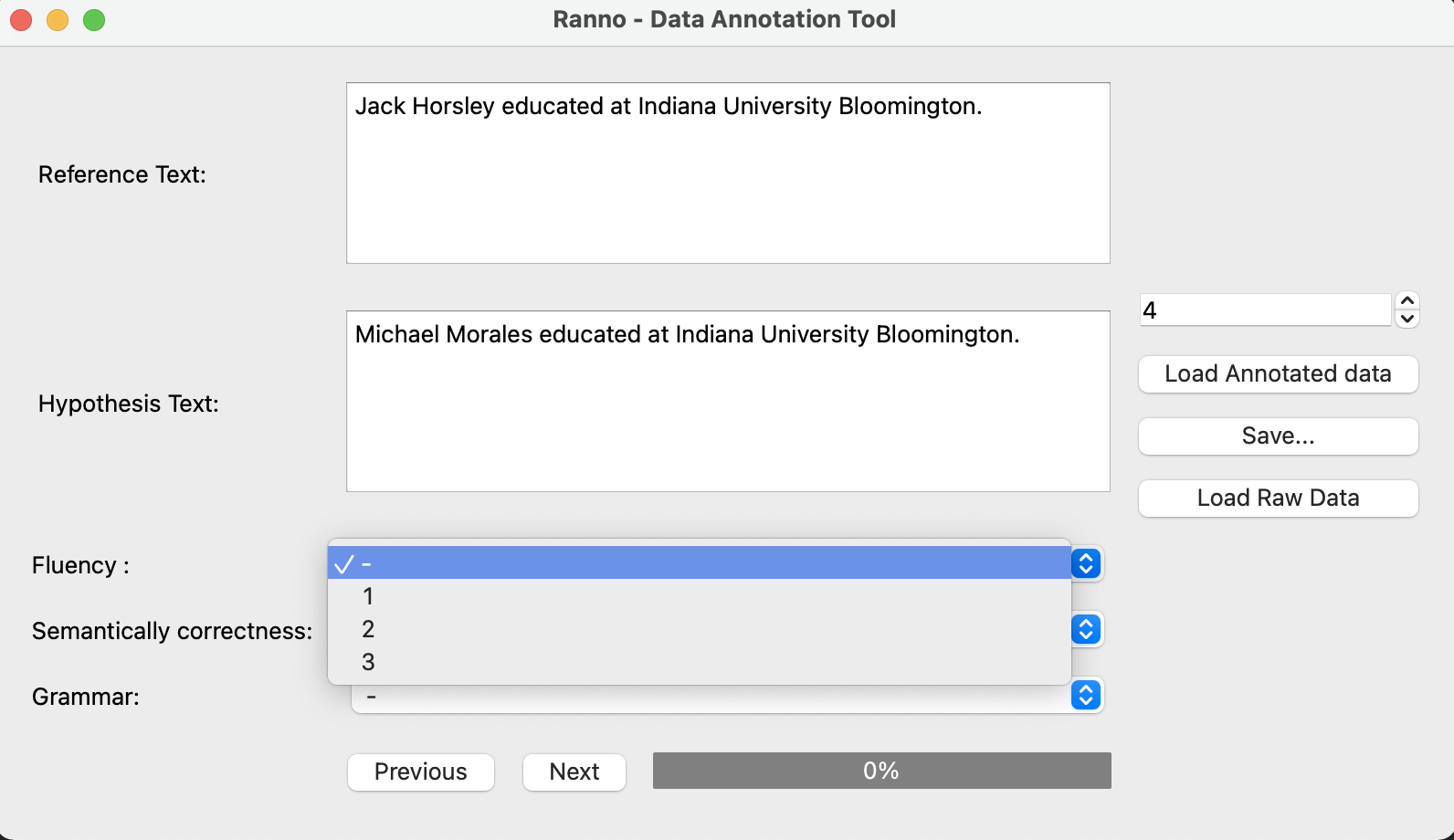}
\caption{The annotation tool used by the annotators.}
    \label{fig:annotool}
\end{figure*}

For all the annotation processes, we use the annotation tool shown in Figure~\ref{fig:annotool}. The tool is developed using Python programming language. Annotators can load their data into the tool in JSON format by selecting the \textit{Load Raw Data} button. An example annotation step is shown in Figure~\ref{fig:annotool}. The reference and hypothesis sentences are displayed in different text windows. The annotators were asked to annotate the data based on \textit{Fluency}, \textit{Semantically correctness} and \textit{Grammar}. Annotators can choose a value on a scale of [1,3] for each category, from the corresponding drop-down option. Finally, the annotated text can be saved for evaluation using the \textit{save} button, which saves the annotated data in JSON format.

\end{document}